\documentclass[runningheads]{llncs}

\usepackage{eccv}

\usepackage{eccvabbrv}

\usepackage{graphicx}
\usepackage{booktabs}
\usepackage{tikz}
\usepackage{multirow}
\usepackage{multicol}
\definecolor{gold}{HTML}{FBF2D2}
\definecolor{silver}{HTML}{DDDDDD}
\definecolor{bronze}{HTML}{EED2B8}

\definecolor{goldD}{HTML}{D9AE13}
\definecolor{silverD}{HTML}{909090}
\definecolor{bronzeD}{HTML}{9A5F26}

\definecolor{catGreen}{HTML}{238763}
\definecolor{catBlue}{HTML}{1F70AE}

\newcommand{\medal}[3]{\tikz[baseline=(char.base)]{\node[rounded corners=2pt,fill=#1,draw=#2,inner sep=1.5pt] (char) {#3};}}

\newcommand{\bm}[2]{
    \ifcase#1\or
      {\medal{gold}{goldD}{\textbf{#2}}}
    \or 
      {\medal{silver}{silverD}{#2}}
    \or 
      {\medal{bronze}{bronzeD}{#2}}
    \else 
      #2
    \fi\ignorespaces
}

\definecolor{trainorange}{HTML}{F7B059}
\definecolor{inferblue}{HTML}{5ADAFA}

\newcommand{\neworrenewcommand}[1]{\providecommand{#1}{}\renewcommand{#1}}
\newcommand{\resultrow}[9]{
\neworrenewcommand{\rresultrow}[3]{#1 & #2 & #3 & #8 & #9 & ##1 & ##2 & #4 & #5 & #6 & #7 & ##3 \\}
\rresultrow
}

\usepackage[accsupp]{axessibility}  

\usepackage{hyperref}

\usepackage{orcidlink}

\begin{document}

\title{TransFusion -- A Transparency-Based Diffusion Model for Anomaly Detection} 

\titlerunning{TransFusion}

\author{Matic Fučka \and
Vitjan Zavrtanik \and
Danijel Skočaj}

\authorrunning{M.~Fučka et al.}

\institute{University of Ljubljana, Faculty of Computer and Information Science, Slovenia
\email{\{matic.fucka, vitjan.zavrtanik, danijel.skocaj\}@fri.uni-lj.si}\\
}

\maketitle

\begin{abstract}
    Surface anomaly detection is a vital component in manufacturing inspection. Current discriminative methods follow a two-stage architecture composed of a reconstructive network followed by a discriminative network that relies on the reconstruction output. Currently used reconstructive networks often produce poor reconstructions that either still contain anomalies or lack details in anomaly-free regions. Discriminative methods are robust to some reconstructive network failures, suggesting that the discriminative network learns a strong normal appearance signal that the reconstructive networks miss. We reformulate the two-stage architecture into a single-stage iterative process that allows the exchange of information between the reconstruction and localization. We propose a novel transparency-based diffusion process where the transparency of anomalous regions is progressively increased, restoring their normal appearance accurately while maintaining the appearance of anomaly-free regions using localization cues of previous steps. We implement the proposed process as TRANSparency DifFUSION (TransFusion), a novel discriminative anomaly detection method that achieves state-of-the-art performance on both the VisA and the MVTec AD datasets, with an image-level AUROC of 98.5\% and 99.2\%, respectively. Code: \textcolor{magenta}{\url{https://github.com/MaticFuc/ECCV_TransFusion}}
\end{abstract}

\section{Introduction}

\begin{figure}[t]
\centering
\includegraphics[width=\columnwidth]{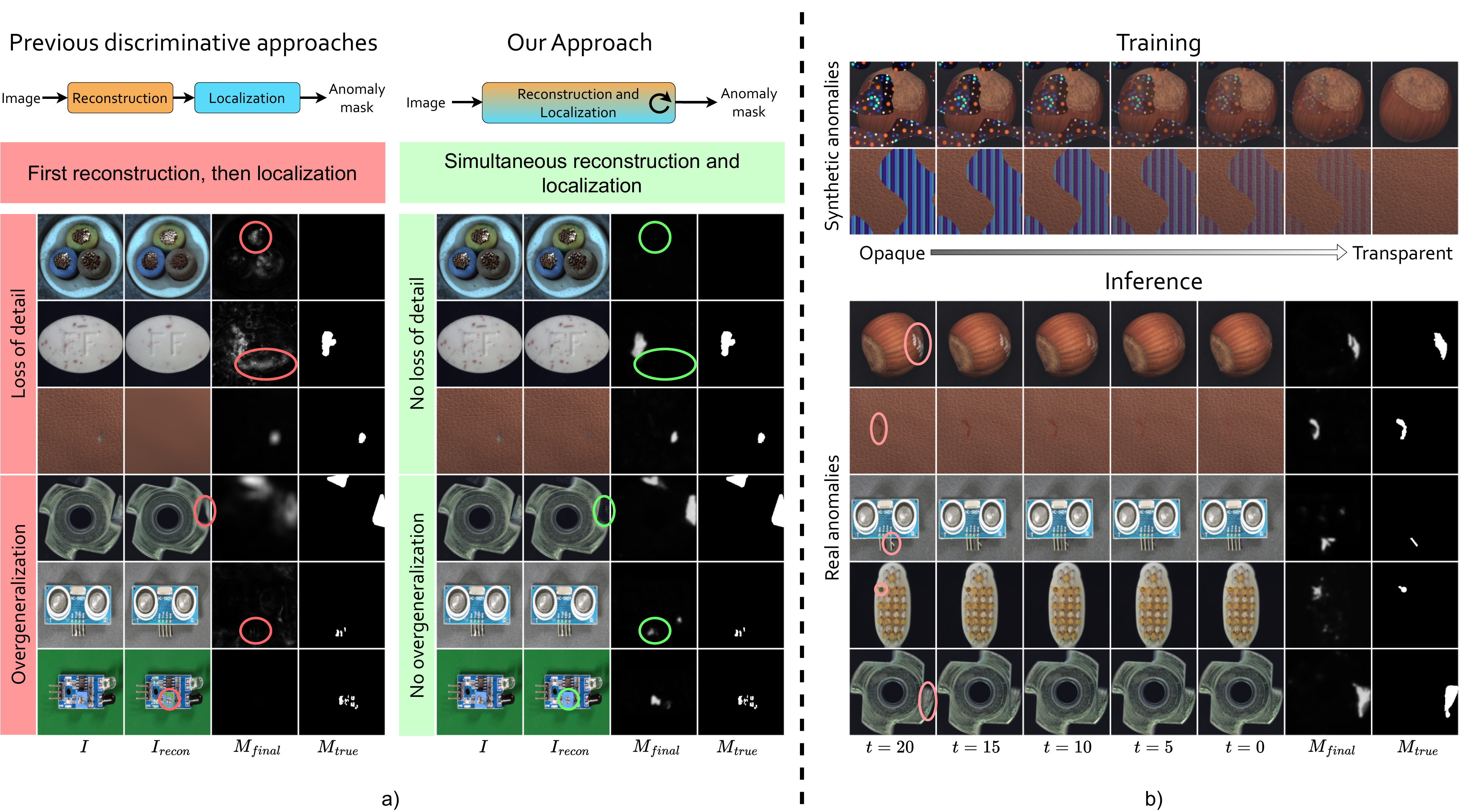}
\caption{a) Different than previous discriminative approaches, the proposed approach simultaneously reconstructs and localizes the anomalies through an iterative process, which results in a more potent normality model capable of detecting harder near-distribution anomalies. b) The reformulated diffusion model iteratively erases the anomalous regions during the reverse process. Training on synthetic anomalies (top) generalizes well to real anomalies (marked with \textcolor{red}{red circles}) seen at inference (bottom), leading to accurate output masks $M_{final}$ that closely match the ground truth $M_{true}$. 
}
\label{fig:transfusion-idea}
\end{figure}

The primary objective of surface anomaly detection is the identification and localization of anomalies in images. In the standard problem setup, only anomaly-free (normal) images are used to learn a normal appearance model and any deviations from the learned model are classified as anomalies. Surface anomaly detection is commonly used in various industrial domains~\cite{mvtec, visa, mvtec-loco} where the limited availability of abnormal images, along with their considerable diversity, makes training supervised models impractical.

Many of the recent surface anomaly detection methods follow the  discriminative~\cite{dsr, draem, simplenet, zhang2023destseg, ldm_draem} paradigm. Discriminative methods are trained to localize simulated anomalies. Discriminative methods typically follow a two-stage architecture: a normal-appearance model followed by a discriminative network. The normal-appearance model learns the anomaly-free object appearance and enables the detection of visual deviations. The discriminative network accurately localizes the anomalies and provides the per-pixel anomaly segmentation mask using the rich signal of the output of the normal-appearance model. Typically, the normal-appearance model is implemented as a reconstructive network.
While the discriminative paradigm hailed the best performances in the past, it started to lag behind with the introduction of more challenging datasets~\cite{visa}.

The failures in reconstructive networks that hurt discriminative methods' downstream anomaly detection capability can be characterised by two core issues. First, reconstructive methods may \textit{overgeneralize}, which causes them to reconstruct even anomalous regions, leading to false negative detections. Second, due to the limited image generation capabilities of the commonly used reconstructive architectures, fine-grained details in normal regions tend to be erased, leading to \textit{loss-of-detail} in normal regions, causing false positive detections. Some samples of these failures can be seen in Figure~\ref{fig:transfusion-idea} a). In an attempt to address these two problems, the autoencoder-based reconstructive network of DR{\AE}M~\cite{draem} has been replaced with a diffusion model in previous works~\cite{ldm_draem}. While the quality of reconstructions was somewhat improved, the loss-of-detail and overgeneralization problems remain in many cases, suggesting that simply replacing the reconstructive subnetwork with a more powerful image generation model is insufficient.

In some cases, discriminative methods can successfully localize an anomaly despite the reconstruction network's failure. This suggests that the discriminative network has the ability to learn the normal appearance signals that the reconstruction network misses. Similarly, discriminative methods can fail to localize an anomaly despite the reconstruction network's success. Interaction between the reconstruction and localization normal-appearance signals might enable the extraction of additional information that is complementary to the information provided by each network and improves downstream anomaly detection performance. This interaction is not done with the current two-stage architecture most discriminative methods follow.

To address the problems of discriminative methods, we propose a novel \textit{transparency-based diffusion process} reformulated explicitly for surface anomaly detection. Through the proposed diffusion process, the transparency of anomalies is iteratively increased so that they are gradually replaced with the corresponding normal appearance (Figure~\ref{fig:transfusion-idea} b), effectively erasing the anomalies. Throughout the proposed process, the anomalies are simultaneously localized and restored to their anomaly-free appearance. This enables a precise anomaly-free reconstruction of the anomalous regions -- addressing \textit{overgeneralization}. Additionally, localization information is used to keep the anomaly-free regions intact -- addressing the \textit{loss-of-detail} problem (Figure~\ref{fig:transfusion-idea} a). To implement the transparency-based diffusion process, we propose \textit{TransFusion (TRANSparency DifFUSION)}, a surface anomaly detection method that integrates the powerful appearance modelling capabilities of diffusion models in the discriminative anomaly detection paradigm. Compared to the previously used reconstructive networks inside discriminative methods that attempted to implicitly detect and restore the anomaly-free appearance of anomalous regions in a single step~\cite{draem,dsr,zhang2023destseg}, TransFusion can maintain more accurate restorations of anomalous regions without the overgeneralization problem and without loss-of-detail in the anomaly-free regions. Due to the iterative nature of the reformulated diffusion process, TransFusion is able to focus on various visual characteristics of anomalies at various time-steps, even potentially addressing the regions previous iterations may have missed. Additionally, the localization information of previous steps can be used as a cue in the reconstruction process, highlighting the potentially anomalous regions. This enables high-fidelity anomaly-free reconstructions and improves the downstream anomaly detection performance significantly, compared to previous discriminative approaches.

The main contributions of our work are as follows:
\begin{itemize}
    \item We propose a novel transparency-based diffusion process reformulated explicitly for the problem of surface anomaly detection. The proposed diffusion process iteratively increases the transparency of anomalies and simultaneously provides their explicit localization.
    \item We propose TransFusion - A strong discriminative anomaly detection model that implements the transparency-based diffusion process. TransFusion directly addresses the overgeneralization and loss-of-detail problems of recent discriminative anomaly detection methods, leading to a strong anomaly detection performance even in difficult near-in-distribution scenarios.
\end{itemize}
We perform extensive experiments on two challenging datasets and show that TransFusion achieves state-of-the-art results in anomaly detection on two standard challenging datasets -- VisA~\cite{visa} and MVTec AD~\cite{mvtec}, with an AUROC of 98.5\% and 99.2\%, respectively. 
TransFusion sets a new state-of-the-art in anomaly detection in terms of the mean across both datasets, achieving a 98.9\% AUROC.

\section{Related Work}

\noindent\textbf{Surface anomaly detection} has been a subject of intense research in recent years, and various approaches have been proposed to address this task. Methods can be divided into three main paradigms: reconstructive, embedding-based, and discriminative.

\textit{Reconstructive methods} train an autoencoder-like network~\cite{ae-ssim, ae-2014, ZavrtanikInpainting} or a generative model~\cite{ganomaly, fanogan, AnnoDDPM, ddpm-noise-anomaly} and assume that anomalies will be poorly reconstructed compared to the normal regions making them distinguishable by reconstruction error. The poor reconstruction assumption does not always hold, leading to poor performance.

\textit{Embedding-based methods} use feature maps~\cite{simplenet, NPad} extracted with a pretrained network to learn normality on these maps. Patchcore~\cite{patchcore} creates a coreset memory bank out of the extracted normal features. Several normalizing-flow-based~\cite{cflow-ad, cs-flow, fastflow, u-flow} approaches have been proposed as well. Some methods utilize a student-teacher~\cite{uninformed, reverse_dist, ast} network and assume that the student will not be able to produce meaningful features for the anomalies as it had not seen them during training. All these methods assume that the distribution of normal regions will be well represented in the training data and fail on rare normal regions unseen during training, producing false positives. 

\textit{Discriminative methods} use synthetically generated defects~\cite{draem,dsr,cutpaste,zhang2023destseg,memseg,simplenet} to train their model with the idea that the model can then generalize on real anomalies. In seminal works of this paradigm, such as DR{\AE}M~\cite{draem}, a two-stage architecture was proposed. First, a reconstructive module is trained to restore the normal appearance, and then, a discriminative network is trained to segment synthetic anomalies. This idea has been followed by the vast majority of models~\cite{memseg,cutpaste,dsr,zhang2023destseg,ldm_draem} inside this paradigm. The normal appearance can also be modelled using pretrained features~\cite{zhang2023destseg,memseg,simplenet}. 
DiffAD~\cite{ldm_draem} has tried to improve DRAEM~\cite{draem} by exchanging the reconstructive subnetwork with a more powerful appearance modelling model, a diffusion model, but the overgeneralization and loss-of-detail problems remained. This suggests that the standard two-stage approach is not optimal for harder near-distribution anomalies.

\begin{figure*}[t]
    \centering
    \includegraphics[width=\textwidth]{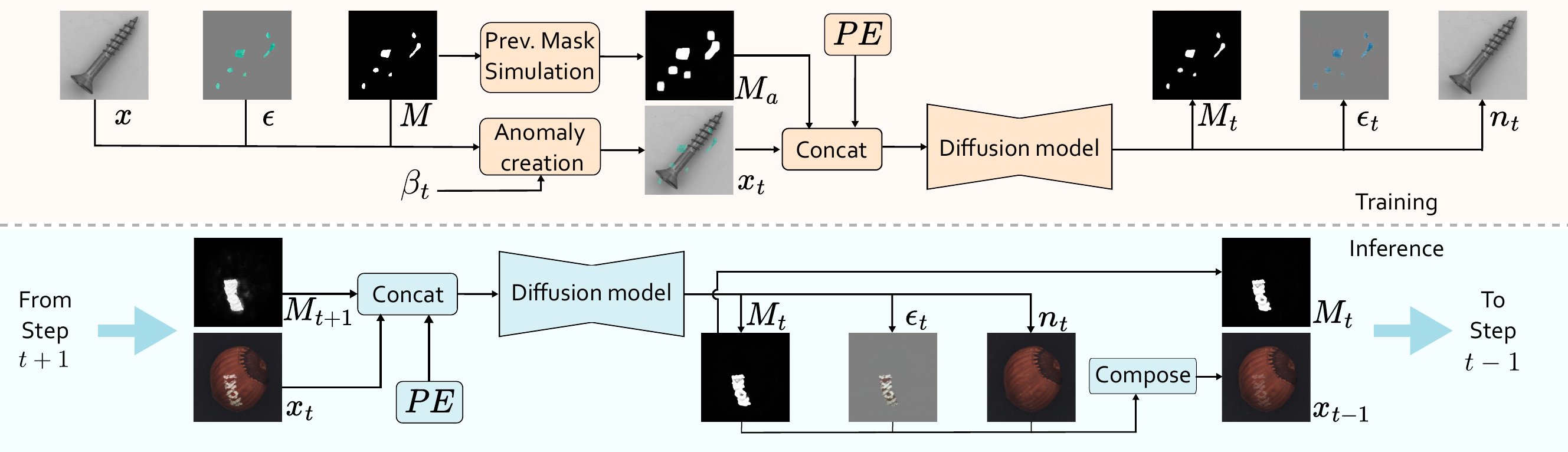}
    \caption{
    TransFusion's \textcolor{trainorange}{training} and \textcolor{inferblue}{inference} pipelines. \textcolor{trainorange}{Training} examples are created from normal images $x$ by generating the anomaly mask $M$ and the anomaly appearance $\epsilon$ and imposing them on $x$ according to the transparency schedule $\beta_t$. The resulting image $x_t$ contains synthetic anomalies. TransFusion is guided by an augmented mask $M_a$. TransFusion outputs the estimated anomaly mask $M_t$, the anomaly appearance $\epsilon_t$, and the normal appearance $n_t$. At \textcolor{inferblue}{inference}, TransFusion infers $M_t$, $\epsilon_t$, and $n_t$ from the input image and constructs the next step image according to Eq. \ref{eq:next_step_eq}. The predicted mask $M_t$ and the constructed $x_{t-1}$ are used as the input in the next step.}

    \label{fig:transfusion}
\end{figure*}

\noindent\textbf{Diffusion models} recently emerged as state-of-the-art in image generation~\cite{ddpm}. They have been extended to various domains, such as audio~\cite{diffusion-audio, diffusion-speech} and text generation~\cite{diffusion-text, diffusion-multinomial-text}. Methods have also been proposed that tackle problems such as semantic segmentation~\cite{segdiff, medsegdiff} and object detection~\cite{diffusiondet}. It has also been shown that the Gaussian noise-based diffusion process is not necessary for all problems~\cite{cold-diffusion, diffusiondet}.

\noindent\textbf{Diffusion-based anomaly detection} Wyatt \textit{et. al.}~\cite{AnnoDDPM} proposed AnoDDPM which is based on a standard diffusion architecture~\cite{ddpm}. AnoDDPM was applied to a medical image dataset and achieved state-of-the-art results. Lu \textit{et. al.}~\cite{ddpm-noise-anomaly} proposed using a DDPM to simultaneously predict the noise and to generate features that mimic the features extracted from a pretrained convolutional neural network. DiffAD~\cite{ldm_draem} exchanged the autoencoder from DR{\AE}M~\cite{draem} with a latent diffusion model to limited success. All recent diffusion approaches face problems with loss-of-detail in the normal regions. As a result, they exhibit a high rate of false positives. This suggests that naively applying the standard diffusion process is insufficient for surface anomaly detection.

\section{TransFusion}

Discriminative anomaly detection approaches attempt to reconstruct the normal visual appearance of anomalies and localize them based on the output of the reconstruction module. An appropriate diffusion model is defined to reformulate this two-stage approach as an iterative one-stage process in order to achieve better detection robustness and reconstruction capability. Previous work~\cite{cold-diffusion} has established that a variety of iterative processes can be used to achieve the desired diffusion effect. In the proposed transparency-based diffusion process reformulation, images are thought of as a composition of anomalous and normal components, partitioned by the anomaly mask $M$. The anomalous regions are expressed as a linear interpolation between the anomalous and the normal appearance at each step to frame the anomaly localisation and restoration as an iterative process. This equates to the transparency of the anomalous regions increasing throughout the diffusion process (Figure~\ref{fig:transfusion-idea} b). In this section, we describe TransFusion in detail.

\subsection{Transparency-based diffusion model}
\label{ch:diff_proc}

In the transparency-based diffusion process reformulation, each image $I$ is expressed as a composition of the normal appearance $N$, the anomaly appearance $A$, the anomaly mask $M$, and the blending factor between the anomalous and the normal appearance $\beta$, i.e., the transparency level of the anomaly:
\begin{equation}
I = \overline{M} \odot N + \beta (M \odot A) + (1-\beta) (M \odot N),
\label{eq:anom_img}
\end{equation}
where $M$ is a binary mask where the anomalous pixels are set to $1$ and $\overline{M}$ is the inverse of $M$. The anomalous region is an interpolation between the anomaly appearance $A$ and the normal appearance $N$ in the region specified by the anomaly mask $M$. The transparency of the anomalous region is defined by $\beta$. The restoration of the normal appearance from an anomalous image $I$ can be modelled as an iterative process of gradually increasing the anomaly transparency until only the normal appearance remains. This is not a trivial task, since the accurate localization $M$, normal appearance $N$, and anomaly appearance $A$ must be inferred from the input image $I$.

\textit{During training}, images containing synthetic anomalies and their corresponding anomaly masks are used.
For each step in the \textit{forward process}, the value of $\beta$ is gradually increased, thus decreasing the transparency of anomalies, and increasing their prominence. Let $x_t$ denote the anomalous image $I$ at time step $t$. The transparency schedule is denoted as $\beta_0 < \beta_1 < ... < \beta_{T-1} < \beta_{T}$, where $\beta_0 = 0$ and $\beta_T = 1$. Eq.~(\ref{eq:anom_img}) is rewritten to correspond to timestep $t$ by substituting the variables $A$ with $\epsilon_t$, $M$ with $M_t$, and $N$ with $n_t$: 
\begin{align}
\begin{split}
x_t &= \overline{M}_t \odot n_t + \beta_t (M_t \odot \epsilon_t) + (1-\beta_t) (M_t \odot n_t).
\label{eq:t}
\end{split}
\end{align}  

The image with more transparent anomalies $x_{t-1}$ at iteration $t-1$ is then computed:
\begin{align}
\begin{split}    
x_{t-1} &= \overline{M}_{t-1} \odot n_{t-1} + \beta_{t-1} (M_{t-1} \odot \epsilon_{t-1}) + (1-\beta_{t-1}) (M_{t-1} \odot n_{t-1}).    
\label{eq:t-1}
\end{split}
\end{align}

$\beta_t$ decreases between steps $t$ and $t-1$, while the correct values of $M_t$, $n_t$ and $\epsilon_t$ are predefined and remain constant throughout the forward process. We can thus write $M_t = M_{t-1} = \ldots = M$, $\epsilon_t = \epsilon_{t-1} = \ldots = A$ and $n_t = n_{t-1} = \ldots = N$. After substituting $M_{t-1}$ for $M_t$, $\epsilon_{t-1}$ for $\epsilon_t$ and $n_{t-1}$ for $n_t$ in Eq.~(\ref{eq:t-1}), subtracting it from Eq.~(\ref{eq:t}) and then rearranging it, the transition between steps $x_t$ and $x_{t-1}$ is computed:
\begin{align}
\begin{split}
x_{t-1} &= x_t - (\beta_t - \beta_{t-1}) (M_t \odot \epsilon_t) + (\beta_t - \beta_{t-1}) (M_t \odot n_t).
\label{eq:next_step_eq}
\end{split}
\end{align}

At each time step in the \textit{reverse process}, the value of $x_t$ moves towards the anomaly-free $x_0$ by an amount influenced by $\beta_t - \beta_{t-1}$. The anomaly's transparency is therefore gradually increased, reconstructing the normal appearance until the final anomaly-free restoration $x_0$ is reached. This requires an accurate estimation of the anomaly mask $M_t$, the normal appearance $n_{t}$ and the anomaly appearance $\epsilon_t$ at each time step.

\subsection{Architecture}
\label{ch:multihead}

The architecture of TransFusion, depicted in Figure~\ref{fig:transfusion}, is based on ResUNet~\cite{resunet}, which is commonly used in diffusion models. TransFusion has three prediction heads, which output the anomaly appearance $\epsilon_t$, anomaly mask $M_t$, and the normal appearance $n_t$, enabling the generation of the image in the next reverse step according to Eq.~(\ref{eq:next_step_eq}). The anomaly and normal appearance heads consist of a single convolutional layer, while the anomaly mask head consists of a BatchNorm, SiLU and a convolutional layer. 

The input to the diffusion model at each timestep consists of four elements: the current reconstruction estimate $x_t$, the mask estimate $M_t$, the 2D sinusoidal positional encoding $PE$~\cite{positional_encoding}, and the timestep $t$. All the elements are channel-wise concatenated except for the timestep embedding, which is added to the features. $PE$ helps the model to learn the global composition of some objects. During training, images containing synthetic anomalies are generated from an anomaly-free image $x$, the anomaly mask $M$, and the anomaly appearance $\epsilon$. The input image $x_t$ is generated according to Eq.~\ref{eq:t}, where $n_t=x$, $\epsilon_t=\epsilon$ and $M_t=M$, and the $\beta$ schedule for the sampled timestep $t$. Losses for the prediction head outputs $n_t$, $M_t$ and $\epsilon_t$ are calculated using $x$, $M$ and $\epsilon$ as ground truth values, respectively.

Separate loss functions are used for each prediction head. The \textit{normal appearance prediction head} uses the structural similarity (SSIM) loss~\cite{ssim} and the $\mathcal{L}_1$ loss:
\begin{equation}
\mathcal{L}_{n} = SSIM(n_t, x) + \mathcal{L}_1(n_t, x).
\end{equation}
The \textit{anomaly mask} head uses the focal loss~\cite{focal} and the Smooth $\mathcal{L}_1$ loss, commonly used in discriminative anomaly detection~\cite{memseg,draem}:
\begin{equation}
\mathcal{L}_{m} = \alpha \mathcal{L}_{foc}(M_t,M) + \mathcal{L}_{1Smooth}(M_t,M).
\end{equation}
The weighting parameter $\alpha$ is set to 5 in all experiments.
The \textit{anomaly appearance prediction head} employs the standard $\mathcal{L}_2$ reconstruction loss:
\begin{equation}
\mathcal{L}_{a} = \mathcal{L}_{2}(\epsilon_t, \epsilon).
\end{equation}

To ensure the consistency between difusion steps, where $x_{t-1}$ is computed from the estimated $M_t$, $\epsilon_t$, $n_t$ and the previous step $x_t$ using Eq.~(\ref{eq:next_step_eq}), an additional \textit{consistency loss} function $\mathcal{L}_{c}$ is employed. $\mathcal{L}_{c}$ compares the predicted $x_{t-1}$ with the ground truth $\tilde{x}_{t-1}$ computed using the ground truth $M$, $\epsilon$, and $x$:
\begin{equation}
\mathcal{L}_{c} = \mathcal{L}_2(x_{t-1}, \tilde{x}_{t-1}).
\end{equation}
The complete TransFusion loss is then given as:
\begin{equation}
\label{eq:overall}
\mathcal{L} = \mathcal{L}_{n} + \mathcal{L}_{m} + \mathcal{L}_{a} + \mathcal{L}_{c}.
\end{equation}

\subsection{Synthetic anomaly generation}
\label{ch:synt_anom}

We directly adopt the synthetic anomaly generation from MemSeg~\cite{memseg}, which is an extension to the synthetic anomaly generation proposed by DR{\AE}M~\cite{draem}. Synthetic anomalies are generated by pasting out-of-distribution regions on the anomaly-free inputs, outputting the image containing synthetic anomalies $I$ and the anomaly mask $M$. $M$ is generated using \textit{Perlin noise}~\cite{perlin1985image}. Synthetic anomalous examples are shown in the top part of Figure~\ref{fig:transfusion-idea} b). Depending on the timestep used, anomalies are generated at different transparency levels.

It would be unrealistic to expect that the current mask estimate $M_t$ would be perfect during inference. To mimic this observation during training, the previous mask estimate imperfection is simulated. The \textit{simulated previous mask estimate} is obtained by thresholding the Perlin noise map used for generating $M$, resulting in a reduction or an expansion of the size of $M_t$. $M_t$ is also dropped during training in $25\%$ of training samples.

\subsection{Inference}
\label{ch:final_mask}

\begin{figure}[t]
    \centering
    \includegraphics[width=0.8\columnwidth]{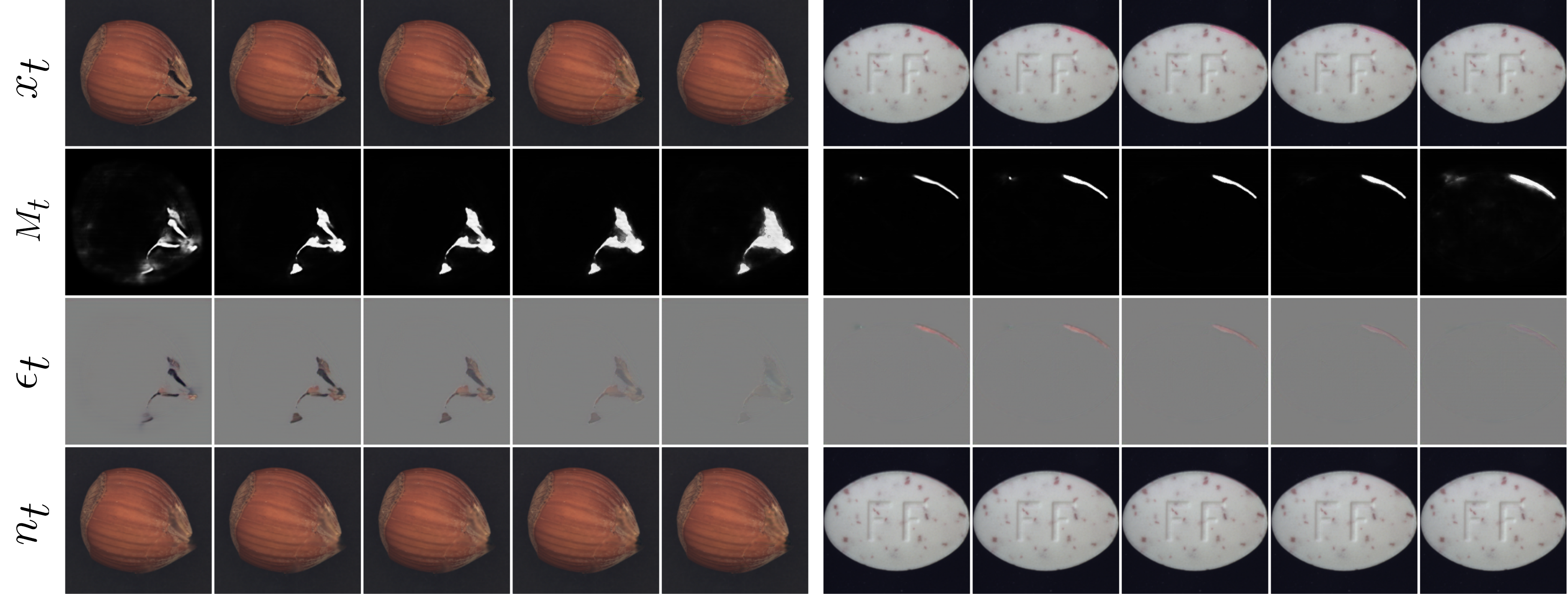}
    \caption{TransFusion inference. For every fourth timestep, the input image $x_t$ and the predictions for the mask $M_t$, anomaly appearance $\epsilon_t$ and normal appearance $n_t$ are shown. As seen in the top row, TransFusion first reconstructs larger anomalies and inpaints the details near the end of the reconstruction process.}
    \label{fig:inference-example}
\end{figure}

At \textit{inference}, Figure~\ref{fig:transfusion}, the starting mask estimate is initialized to all zero values.
Then, the reverse process of T time steps is performed. T is set to 20 in all experiments unless stated otherwise. At each time step $t$, the current approximation of the reconstructed image $x_t$ is channel-wise concatenated with the binarized previous mask estimate $M_{t+1}$ and positional encoding $PE$. This composite input and the current timestep $t$ are fed into the diffusion model. The model's output consists of the current mask estimate $M_t$, an anomaly appearance estimation $\epsilon_t$, and a normal appearance estimation $n_t$ (Figure~\ref{fig:transfusion}, bottom middle). Based on these outputs, the next step  $x_{t-1}$ is predicted using Eq.~(\ref{eq:next_step_eq}) (Figure~\ref{fig:transfusion}, bottom right). Anomaly mask $M_t$ is binarized by thresholding and used in the next step. An example of the inference process is visualized in Figure~\ref{fig:inference-example}. The reverse process iteratively reduces the transparency of the anomalous regions, progressively restoring the anomaly-free appearance of the image. At time step 0, the result is a fully reconstructed anomaly-free image $x_0$.

The \textit{final anomaly mask} $M_{final}$ is derived from $M_{disc}$, the pixel-wise mean of anomaly masks $M_t$, with $t$ going from $1$ to $T$, produced throughout the reverse process and from $M_{recon}$, the reconstruction error between the initial image $x$ and the diffusion model output $x_0$.

To obtain the final mask $M_{final}$, a weighted combination of $M_{disc}$ and $M_{recon}$ is performed:
\begin{equation}
\label{eq:final_mask_calc}
M_{final} =(\lambda M_{disc} + (1 - \lambda) M_{recon}) * f_{n},   
\end{equation}
where the influence of $M_{disc}$ and $M_{recon}$ is weighted by $\lambda$ ($\lambda$=0.95 in all experiments), $f_n$ is a mean filter of size $n \times n$ (in our case $7\times7$) and $*$ is the convolution operator. The mean filter smoothing is performed to aggregate the local anomaly map responses for a robust image-level score estimation. The image-level anomaly score $AS$ is obtained by the maximum value of $M_{final}$:
\begin{equation}
 AS = max(M_{final}).
 \label{eq:anomaly_score}
\end{equation}

Including both $M_{disc}$ and $M_{recon}$ gives the final mask $M_{final}$ a balanced anomaly representation, allowing it to benefit from both discriminative and reconstructive cues.

\section{Experiments}

\subsection{Datasets}

Experiments are performed on two standard anomaly detection datasets: the VisA dataset~\cite{visa} and the MVTec AD dataset~\cite{mvtec}. The VisA dataset is comprised of 10,821 images distributed across 12 object categories, while the MVTec AD dataset contains 5,354 images encompassing 5 texture categories and 10 object categories. Notably, both datasets provide pixel-level annotations for the test images, enabling accurate evaluation and analysis.

\subsection{Evaluation metrics}

Standard anomaly detection evaluation metrics are used. The image-level anomaly detection performance is evaluated by the Area Under the Receiver Operator Curve (AUROC), while for the pixel-level anomaly localization the Area Under the Per Region Overlap (AUPRO) is utilized.

\subsection{Implementation details}
During both training and inference, 20 steps ($T=20$) are used in the diffusion process with a linear transparency ($\beta$) schedule ranging from 0 to 1. 
The model was trained for 1500 epochs using the AdamW optimizer with a batch size of 8. The learning rate was set to $10^{-5}$ and was multiplied by $0.1$ after 800 epochs. Synthetic anomalies were added to half of the training batch. Rotation augmentation was used following DR{\AE}M~\cite{draem}. A standard preprocessing approach is employed to ensure experimental consistency. Each image is resized to dimensions of $256\times256$ and subsequently center-cropped to $224\times224$ following recent literature~\cite{patchcore, memseg, simplenet}. The image is then linearly scaled between -1 and 1 following recent diffusion model literature~\cite{ddpm,ddim}. Following the standard protocol for unsupervised anomaly detection, a separate model was trained for each category, and the same hyperparameters were set across both datasets and all categories.

\subsection{Experimental results}

\textit{Anomaly detection} results on VisA are shown in Table~\ref{tb:visa_det_results}. TransFusion achieves the best results on 5 out of the 12 categories and outperforms the previous best state-of-the-art method by $0.4$ percentage points in terms of the mean AUROC performance. On the MVTec AD dataset, TransFusion achieves state-of-the-art results with a mean anomaly detection AUROC of 99.2\%. Results are shown in Table~\ref{tb:mvtec_det_results}.

Due to the significant differences in anomaly types between the VisA and MVTec AD datasets, very few recent methods exhibit the generalization capability necessary to achieve top results for both datasets. Table~\ref{tb:both_det_results} shows results on both VisA and MVTec AD. Additionally, the average scores across both datasets are shown. TransFusion outperforms all recent methods in terms of the average anomaly detection AUROC by $0.3$ percentage points and, more notably, outperforms the next best discriminative method by a significant margin of $4.0$ percentage points, reducing the error by $78.5\%$.

TransFusion also achieves the second highest score in \textit{anomaly localization} when averaged across both datasets, achieving an AUPRO of 91.6\%. 
In terms of anomaly detection, TransFusion outperforms competing methods significantly on the VisA dataset and achieves state-of-the-art performance on MVTec AD. TransFusion also outperforms other diffusion-based methods, AnoDDPM~\cite{AnnoDDPM}, DiffAD~\cite{ldm_draem} and AnomDiff~\cite{ddpm-noise-anomaly}, by a significant margin, which suggests that simply relying on a standard diffusion process for reconstruction may not be sufficient for anomaly detection. TransFusion also significantly outperforms previous state-of-the-art discriminative methods, such as DR{\AE}M~\cite{draem}, DSR~\cite{dsr}, DiffAD~\cite{ldm_draem} and SimpleNet~\cite{simplenet}, on the VisA dataset in terms of anomaly detection. This suggests that simultaneous localization and reconstruction provide a more potent normality model in comparison to the previous two-stage paradigm. 

\begin{table*}[t]
\centering
\setlength{\tabcolsep}{3pt}
\resizebox{\textwidth}{!}{
\begin{tabular}{lccccccccccc}
\toprule 
\resultrow{\multirow{2}{*}{\textbf{Method}}}{AnoDDPM}{AnomDiff}{\textbf{DiffAD}}{\textbf{DR{\AE}M}}{\textbf{DSR}}{\textbf{SimpleNet}}{PatchCore}{RD4AD}{AST}{EfficientAD}{\textit{TransFusion}}
\resultrow{}{\cite{AnnoDDPM}}{\cite{ddpm-noise-anomaly}}{\cite{ldm_draem}}{\cite{draem}}{\cite{dsr}}{\cite{simplenet}}{\cite{patchcore}}{\cite{reverse_dist}}{\cite{ast}}{\cite{efficientad}}{-}
\midrule 
\resultrow{Candle}{64.9}{64.9}{90.4}{94.4}{\bm2{98.8 }}{95.6}{98.1}{92.2}{\bm1{99.4 }}{98.4}{\bm3{98.3 }}
\resultrow{Capsules}{76.5}{80.0}{87.6}{76.3}{\bm2{99.1 }}{76.7}{85.7}{\bm3{90.1 }}{85.4}{93.5}{\bm1{99.6 }}
\resultrow{Cashew}{94.4}{90.9}{81.4}{90.7}{\bm3{97.6 }}{91.7}{\bm2{98.5 }}{\bm1{99.6 }}{95.1}{97.2}{93.7}
\resultrow{Chewing gum}{91.3}{98.1}{94.0}{94.2}{93.8}{99.1}{99.0}{\bm3{99.7 }}{\bm1{100 }}{\bm2{99.9 }}{99.6}
\resultrow{Fryum}{81.5}{89.2}{87.1}{97.4}{82.9}{95.3}{97.2}{96.6}{\bm1{99.1 }}{96.5}{\bm3{98.3 }}
\resultrow{Macaroni1}{58.8}{77.8}{87.6}{95.0}{87.3}{90.8}{95.7}{\bm2{98.4 }}{93.9}{\bm1{99.4 }}{\bm1{99.4 }}
\resultrow{Macaroni2}{74.5}{61.0}{90.7}{96.2}{83.4}{65.2}{78.1}{\bm1{97.6 }}{72.1}{\bm2{96.7 }}{\bm3{96.5 }}
\resultrow{PCB1}{42.1}{86.7}{75.0}{54.8}{90.5}{60.1}{98.3}{97.6}{\bm1{99.2 }}{\bm3{98.5 }}{\bm2{98.9 }}
\resultrow{PCB2}{90.7}{76.5}{94.6}{77.8}{96.6}{93.3}{97.2}{91.1}{\bm3{98.4 }}{\bm2{99.5 }}{\bm1{99.7 }}
\resultrow{PCB3}{92.3}{80.4}{94.7}{94.5}{94.8}{94.9}{96.2}{95.5}{\bm3{97.4 }}{\bm2{98.9 }}{\bm1{99.2 }}
\resultrow{PCB4}{98.3}{93.8}{97.7}{93.4}{93.5}{98.2}{\bm3{99.0 }}{96.5}{\bm1{99.6 }}{98.9}{\bm1{99.6 }}
\resultrow{Pipe fryum}{72.5}{89.4}{92.7}{\bm3{99.4 }}{97.5}{93.3}{99.4}{97.0}{\bm3{99.4 }}{\bm1{99.6 }}{\bm1{99.6 }}
\midrule   
\resultrow{\textit{Average}}{78.2}{83.7}{89.5}{88.7}{91.6}{87.9}{94.3}{\bm3{96.0 }}{94.9 }{ \bm2{98.1 }}{\bm1{98.5 }}
\bottomrule
\end{tabular}
}
\caption{Comparison of TransFusion in anomaly detection (AUROC) with SOTA on VisA. \textcolor{goldD}{First}, \textcolor{silverD}{second} and \textcolor{bronzeD}{third} place are marked. The names of all previous discriminative approaches are typeset in \textbf{bold}.}
\label{tb:visa_det_results}
\end{table*}

\begin{table*}[t]
\setlength{\tabcolsep}{3pt}
\centering
\resizebox{\textwidth}{!}{
\begin{tabular}{lccccccccccc}
\toprule
\resultrow{\multirow{2}{*}{\textbf{Method}}}{AnoDDPM}{AnomDiff}{\textbf{DiffAD}}{\textbf{DR{\AE}M}}{\textbf{DSR}}{\textbf{SimpleNet}}{PatchCore}{RD4AD}{AST}{EfficientAD}{\textit{TransFusion}}
\resultrow{}{\cite{AnnoDDPM}}{\cite{ddpm-noise-anomaly}}{\cite{ldm_draem}}{\cite{draem}}{\cite{dsr}}{\cite{simplenet}}{\cite{patchcore}}{\cite{reverse_dist}}{\cite{ast}}{\cite{efficientad}}{-}
\midrule 
\resultrow{Carpet}{93.5}{\bm2{99.9 }}{98.3}{97.0}{\bm1{100 }}{97.5}{98.7}{95.3}{99.1}{\bm3{99.3 }}{99.2}
\resultrow{Grid}{93.8}{99.7}{\bm1{100 }}{99.9}{\bm1{100 }}{99.1}{98.2}{\bm1{100 }}{98.7}{99.9}{\bm1{100 }}
\resultrow{Leather}{99.5}{\bm1{100}}{\bm1{100 }}{\bm1{100 }}{\bm1{100 }}{\bm1{100 }}{\bm1{100 }}{97.1}{\bm1{100 }}{\bm1{100}}{\bm1{100 }}
\resultrow{Tile}{99.4}{98.0}{\bm1{100 }}{99.6}{\bm1{100 }}{\bm1{100 }}{98.7}{99.3}{99.1}{99.9}{99.8}
\resultrow{Wood}{99.0}{98.1}{\bm1{100 }}{99.1}{96.3}{\bm1{100 }}{99.2}{99.2}{99.2}{\bm1{100 }}{99.4}
\midrule  
\resultrow{Bottle}{98.4}{99.3}{\bm1{100 }}{99.2}{\bm1{100 }}{\bm1{100 }}{\bm1{100 }}{\bm1{100 }}{\bm1{100 }}{99.9}{\bm1{100 }}
\resultrow{Cable}{52.7}{91.2}{94.6}{91.8}{93.8}{\bm1{99.9 }}{\bm3{99.5 }}{95.0}{\bm3{98.5 }}{95.2}{97.9}
\resultrow{Capsule}{89.0}{84.1}{97.5}{\bm2{98.5 }}{98.1}{97.7}{98.1}{96.3}{\bm1{99.7 }}{97.9}{\bm2{98.5 }}
\resultrow{Hazelnut}{84.5}{97.9}{\bm1{100 }}{\bm1{100 }}{95.6}{\bm1{100 }}{99.9}{\bm1{100 }}{\bm1{100 }}{99.4}{\bm1{100 }}
\resultrow{Metal nut}{92.8}{99.2}{99.5}{98.7}{98.5}{\bm1{100 }}{\bm1{100 }}{\bm1{100 }}{98.5}{99.6}{\bm1{100 }}
\resultrow{Pill}{80.9}{64.7}{97.7}{\bm2{98.9 }}{97.5}{\bm1{99.0 }}{96.6}{96.6}{99.1}{\bm3{98.6 }}{98.3}
\resultrow{Screw}{20.3}{89.9}{97.2}{93.9}{96.2}{\bm2{98.2}}{\bm3{98.1 }}{97.0}{\bm1{99.7 }}{96.9}{97.2}
\resultrow{Toothbrush}{86.4}{96.9}{\bm1{100 }}{\bm1{100 }}{99.7}{99.7}{\bm1{100 }}{90.8}{96.6}{\bm1{100}}{\bm1{100 }}
\resultrow{Transistor}{65.0}{92.3}{96.1}{93.1}{97.8}{\bm1{100 }}{\bm1{100 }}{96.7}{99.3}{\bm2{99.9 }}{98.3}
\resultrow{Zipper}{98.2}{85.5}{\bm1{100 }}{\bm1{100 }}{\bm1{100 }}{99.5}{99.4}{98.5}{99.1}{99.7}{\bm1{100 }}
\midrule    
\resultrow{\textit{Average}}{83.5}{93.1}{98.7}{98.0}{98.2}{\bm1{99.6 }}{99.1}{98.5}{\bm2{99.2 }}{99.1}{\bm2{99.2 }}
\bottomrule
\end{tabular}
}
\caption{Comparison of TransFusion in anomaly detection (AUROC) with SOTA on MVTec AD.
}
\label{tb:mvtec_det_results}

\end{table*}

\begin{table}[h!]
\setlength{\tabcolsep}{3pt}
\centering
\resizebox{0.6\columnwidth}{!}{
\begin{tabular}{l|c|c|cc|cc|cc}
\hline \multirow{2}{*}{\textbf{Method}} & \multirow{2}{*}{\textbf{Venue}} & \multirow{2}{*}{\textbf{Disc.}} & \multicolumn{2}{|c|}{VisA} & \multicolumn{2}{|c|}{MVTec AD} & \multicolumn{2}{|c}{\textit{Average}} \\
& & & Det. & Loc. & Det. & Loc. & Det. & Loc.\\  \hline
AnoDDPM & CVPRW'22 & & 78.2 & 60.5 & 83.5 & 50.7 & 80.9 & 55.6\\
DR{\AE}M & ICCV'21 & \checkmark & 88.7 & 73.1 & 98.0 & 92.8 & 93.3 & 83.0\\
SimpleNet & CVPR'23 & \checkmark & 87.9 & 68.9 & \bm1{99.6 } & 89.6 & 93.8 & 79.3 \\
DiffAD & ICCV'23 & \checkmark & 89.5 & 71.2 & 98.7 & 84.8 & 94.1 & 78.0\\
DSR & ECCV'22 & \checkmark & 91.6 & 68.1 & 98.2 & 90.8 & 94.9 & 79.5\\
Patchcore & CVPR'22 & & 94.3 & 79.7 & 99.1 & 92.7 & 97.0 & \bm3{86.2 }\\
AST & WACV'23 & & 94.9 & \bm3{81.5 } & \bm2{99.2 } & 81.2 & 97.1 & 81.4\\ 
RD4AD & CVPR'22 & & \bm3{96.0 } & 70.9 & 98.5 & \bm2{93.9 } & \bm3{97.3 } & 82.4 \\
EfficientAD & WACV'24 & & \bm2{98.1 } & \bm1{94.0 } & 99.1 & \bm3{93.5 } & \bm2{98.6 } & \bm1{93.7 } \\
\textit{TransFusion} & ECCV'24 & \checkmark & \bm1{98.5 } & \bm2{88.8 } &  \bm2{99.2 } & \bm1{94.3 } & \bm1{98.9 } & \bm2{91.6 } \\ \hline
\end{tabular}
}
\caption{Results in anomaly detection (AUROC) and anomaly localization (AUPRO) on both VisA and MVTec AD. 
}
\label{tb:both_det_results}
\end{table}

\begin{figure*}[t]
    \centering
    \includegraphics[width=\textwidth]{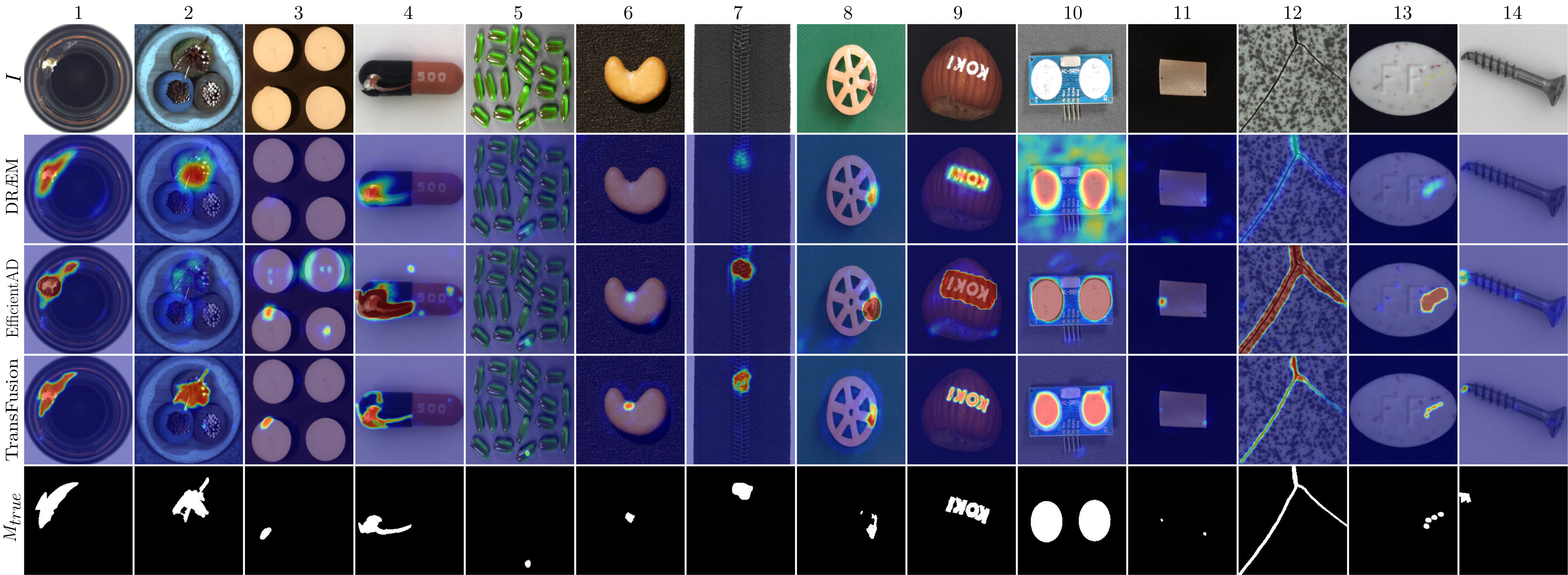}
    \caption{Qualitative comparison of the masks produced by TransFusion and three other state-of-the-art methods. The anomalous images are shown in the first row. The middle four rows show the anomaly mask generated by DR{\AE}M~\cite{draem}, EfficientAD~\cite{efficientad} and TransFusion, respectively. The last row shows the ground truth anomaly mask.}
    \label{fig:mask-comp}
\end{figure*}

\begin{figure}[h!]
    \centering
    \includegraphics[width=\columnwidth]{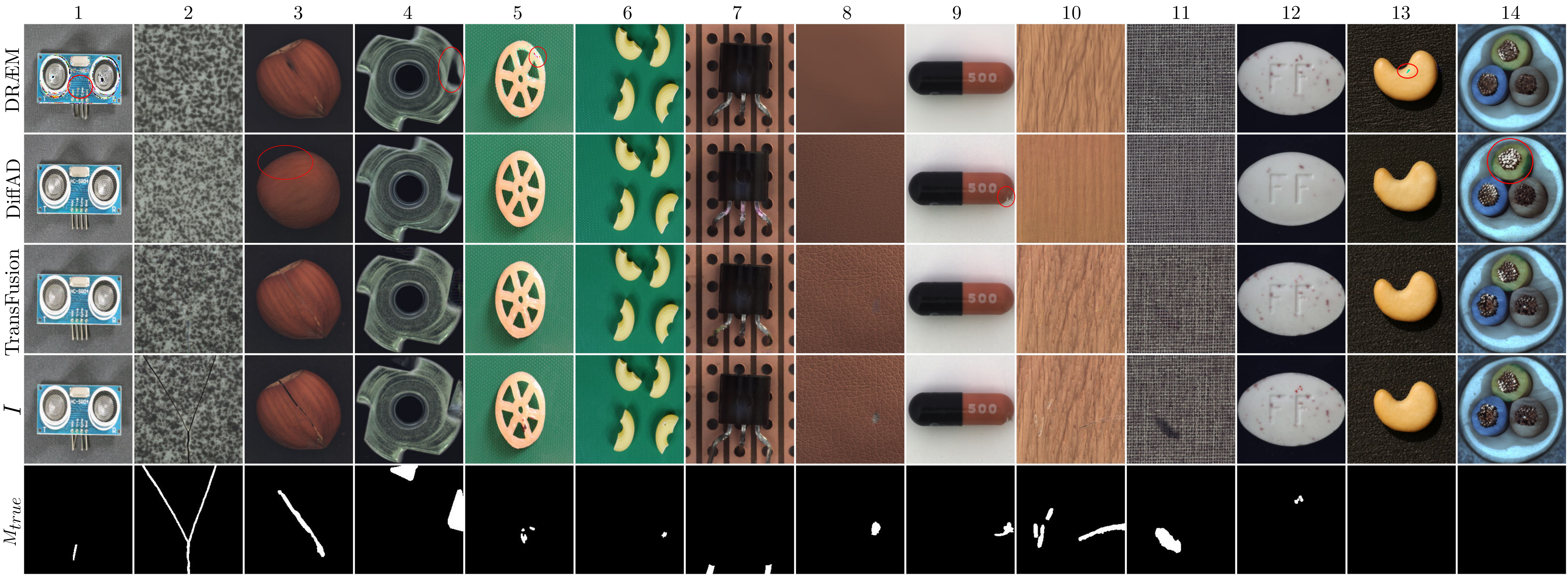}
    \caption{Qualitative reconstruction results. TransFusion better restores anomalies to their normal appearance and better preserves the details in the normal regions than competing methods DR{\AE}M~\cite{draem} and DiffAD~\cite{ldm_draem}. A few of the larger differences are highlighted in red.}
    \label{fig:recon-comp}
\end{figure}

\subsection{Qualitative comparisons}

A qualitative comparison with the state-of-the-art methods DR{\AE}M~\cite{draem} and EfficientAD~\cite{efficientad} can be seen in Figure~\ref{fig:mask-comp}. Note that TransFusion outputs very precise anomaly masks and does not produce significant false positives in the background as opposed to other state-of-the-art methods (Columns 3, 4, 14). Due to being a discriminative network, TransFusion outputs masks (Columns 1-14) that are much sharper than those of EfficientAD, which outputs a feature reconstruction error. DR{\AE}M is unable to accurately detect small near-in-distribution anomalies (Columns 3, 5, 6, 14) mostly present in the VisA~\cite{visa} Dataset. We hypothesize that this is due to the problems encountered in two-stage discriminative approaches, such as DR{\AE}M.

TranFusion exhibits a strong reconstructive ability. A qualitative comparison can be seen in Figure~\ref{fig:recon-comp}. Compared to DR{\AE}M~\cite{draem}, TransFusion outputs higher-quality reconstructions and even produces realistic results in difficult reconstruction cases, such as strong deformations, while maintaining fine-grained details in normal regions. TransFusion better addresses the loss-of-detail problem than the previously proposed method DiffAD~\cite{ldm_draem}. The reconstructions suggest that simultaneous reconstruction and localization produce more powerful normal-appearance models.

\subsection{Ablation study}

\begin{table}[t]
\setlength{\tabcolsep}{3pt}
\centering
\resizebox{0.5\columnwidth}{!}{
\begin{tabular}{llcccc} \hline
                 \multirow{2}{*}{\textbf{Group}} & \multirow{2}{*}{\textbf{Condition}}  & \multicolumn{2}{c}{VisA} & \multicolumn{2}{c}{MVTec AD} \\
                  &   & Det. & Loc. & Det. & Loc. \\ \hline
&  w/o PE & \textcolor{blue}{-1.4} & \textcolor{gray}{-0.1} & \textcolor{gray}{-1.9} & \textcolor{gray}{-3.2} \\
\multirow{-2}{*}{\textit{Input strategies}}& w/o Simulated Mask & \textcolor{gray}{-1.2} & \textcolor{blue}{-8.6} & \textcolor{blue}{-2.4} & \textcolor{blue}{-6.2} \\ \hline
& w/o $\mathcal{L}_{n}$ & \textcolor{blue}{-31.8} & \textcolor{blue}{-45.3} & \textcolor{blue}{-26.0} & \textcolor{blue}{-43.2}\\
& w/o $\mathcal{L}_{m}$ & \textcolor{gray}{-1.5} & \textcolor{gray}{-3.2} & \textcolor{gray}{-2.2} & \textcolor{gray}{-2.7}\\
 & w/o $\mathcal{L}_{a}$ & \textcolor{gray}{-1.1} & \textcolor{gray}{-1.5} & \textcolor{gray}{-1.0} & \textcolor{gray}{-1.0}\\ 
 \multirow{-4}{*}{\textit{Loss Function}} &  w/o $\mathcal{L}_{c}$ & \textcolor{gray}{-0.9} & \textcolor{gray}{-0.3} & \textcolor{gray}{-0.7} & \textcolor{gray}{-0.8}\\ \hline
&  Only $M_1$ & \textcolor{blue}{-1.5} & \textcolor{gray}{-0.3} & \textcolor{blue}{-1.4} & \textcolor{gray}{-2.6}\\
&  Only $M_{disc}$ & \textcolor{gray}{-0.1}  & \textcolor{gray}{+0.1} & \textcolor{gray}{-0.2} & \textcolor{gray}{+0.1}\\[-1pt]
 \multirow{-3}{*}{\textit{Final mask calc.}} &  Only $M_{recon}$ & \textcolor{gray}{-0.9} & \textcolor{blue}{-13.5} & \textcolor{gray}{-0.7} & \textcolor{blue}{-5.9}\\[-1pt] \hline
 &  5 steps & \textcolor{blue}{-1.0} & \textcolor{blue}{-4.3} & \textcolor{blue}{-0.7} & \textcolor{blue}{-1.2}\\
 &  10 steps & \textcolor{gray}{-0.5} & \textcolor{gray}{-1.1} & \textcolor{gray}{-0.7} & \textcolor{gray}{-1.0}\\
\multirow{-3}{*}{\textit{Diffusion step num.}} & 50 steps & \textcolor{gray}{-0.3}  & \textcolor{gray}{+0.7} & \textcolor{gray}{-0.5} & \textcolor{gray}{-0.8}\\[-1pt] \hline
&  Quadratic & \textcolor{blue}{-2.0} & \textcolor{gray}{+1.2} & \textcolor{blue}{-2.0} & \textcolor{gray}{+0.4}\\[-1pt]
\multirow{-2}{*}{\textit{Transparency sched.}}&  Root & \textcolor{gray}{-1.7}  & \textcolor{blue}{-2.7} & \textcolor{gray}{-0.7} & \textcolor{blue}{-1.6}\\ \hline

\textit{TransFusion} & Linear, 20 steps & 98.5 & 88.8 & 99.2 & 94.3\\ \hline   
\end{tabular}
}
\caption{Ablation study results. Detection results are reported in AUROC and localization results are reported in AUPRO. In each row, the difference to the actual model is shown. The highest discrepancy for each experiment group is marked in \textcolor{blue}{blue}.}
\label{tb:ablation}
\end{table}

The results of the evaluation of individual components of TransFusion and it's training process are shown in Table~\ref{tb:ablation}.

\noindent\textbf{Input strategies.} In addition to the image $x_t$, the Positional Encoding (PE) and the simulated previous mask are input during training. The impact of PE and the simulated previous mask is evaluated by excluding each individually from the architecture. \textit{Excluding PE} leads to a $1.4$ percentage points (p.\ p.) drop on VisA and a $1.9$ p.\ p.\ drop on MVTec AD. \textit{Excluding the simulated previous mask} leads to a $1.2$ p.\ p. drop on VisA and a $2.4$ p. p. drop on MVTec AD, showing the benefit of the localization information gained from the previous step. There is also a significant drop ($8.6$ p.\ p.\ on VisA and $6.2$ p.\ p.\ on MVTec AD) in localization when excluding the simulated mask, highlighting its importance for precise localization.

\noindent\textbf{Importance of loss functions.} The importance of each loss function was evaluated by excluding one loss function at a time and training the model. Removing $L_a$, $L_{c}$ or $L_{m}$ reduces the overall anomaly detection performance by approximately $1$ p.\ p.\ on VisA and MVTec AD, demonstrating their usefulness. Notably, removing  $\mathcal{L}_{n}$ leads to a major drop in performance ($31.8$ p.\ p.\ AUROC on VisA, $26$ p.\ p.\ AUROC on MVTec AD), showing the necessity of learning a strong normal appearance model of the object. Without $\mathcal{L}_{n}$, TransFusion may focus on learning the synthetic anomaly appearance, leading to poor generalization. 

\noindent\textbf{Final mask calculation.} The anomaly mask calculation methods using either only the last mask estimate $M_1$, the discriminative mask $M_{disc}$, or the reconstruction mask $M_{recon}$ are evaluated. Using only $M_{recon}$ leads to a $0.9$ p.\ p.\ drop on VisA and a $0.7$ p.\ p.\ MVTec AD in terms of AUROC. $M_{disc}$ can accurately localize the anomalies even without $M_{recon}$, leading to only a $0.1$ and $0.2$ p.\ p.\ drop on VisA and MVTec AD, respectively. The impact of mask averaging throughout the diffusion process is significant since using only the last estimated mask (Last Mask Est.) causes a $1.5$ and $1.4$ p.\ p.\ drop in anomaly detection performance on the VisA and MVTec AD, respectively.

\noindent\textbf{Number of diffusion steps.} The impact of the number of diffusion steps on the anomaly detection performance is evaluated. Although a lower number of steps leads to a poorer normal appearance restoration, TransFusion remains robust across various time-step settings, achieving similar results across both VisA and MVTec AD, even achieving state-of-the-art results on VisA at only 5 timesteps. A higher number of diffusion steps also increases the result in localization on VisA.

\noindent\textbf{Transparency schedule.} The impact of replacing the linear $\beta$ schedule with alternative schedules is evaluated. The Root and the Quadratic schedule are examined, where the $\beta$ values change from $0$ to $1$ using a quadratic or a square-root function, respectively. Using a Quadratic schedule causes a $2$ p.\ p.\ drop in performance on both VisA and MVTec AD. The Root schedule leads to a $1.7$ and a $0.7$ p.\ p.\ drop on the VisA and the MVTec AD, respectively. Interestingly, using a quadratic schedule improves anomaly localization by a $1.2$ p.\ p.\ on VisA and a $0.4$ p.\ p.\ on MVTec AD.

\noindent\textbf{Inference efficiency.} Inference times of various methods can be seen in Table~\ref{tb:inf_speed}. Due to the complexity of diffusion models, TransFusion is slower than some competing methods however, it is faster than other diffusion-based methods. Additionally, reducing the number of inference steps does not drastically reduce performance (Table~\ref{tb:ablation}). Speeding up diffusion models is an active field~\cite{diff_distillation, progressive_diffusion_distillation, consistency_model} and may be helpful to increase the inference speed of TransFusion in the future.

\begin{table}[t]
\centering
\resizebox{0.5\columnwidth}{!}{
\begin{tabular}{lcccc} \toprule
\textbf{Method} & DR{\AE}M~\cite{draem} & Patchcore~\cite{patchcore} & DiffAD~\cite{ldm_draem} & \textit{TransFusion} \\ \midrule
Inference [s] & \bm1{0.05 } & \bm2{0.22 } & 1.00 & \bm3{0.34 } \\ \bottomrule
\end{tabular}
}
\caption{Results for average inference time of a single sample with NVIDIA A100 GPU. Inference times are reported in seconds.}
\label{tb:inf_speed}
\end{table}

\section{Conclusion}
A novel, transparency-based diffusion process is proposed, where the transparency of the anomalous regions is gradually increased, effectively removing them and restoring their normal appearance. TransFusion, a novel discriminative anomaly detection method that implements the transparency-based diffusion process, is proposed. With simultaneous localization and reconstruction, TransFusion is able to produce accurate anomaly-free reconstructions of anomalies while maintaining the appearance of normal regions, thus addressing both the overgeneralization and loss-of-detail problems of commonly used reconstructive models inside discriminative approaches. TransFusion achieves state-of-the-art results in anomaly detection on the standard VisA and MVTec AD datasets, achieving an AUROC of 98.5\% and 99.2\% for both datasets, respectively. 
The versatility of TransFusion and its robustness to near-in-distribution anomalies are further validated by the state-of-the-art performance across both datasets, where TransFusion achieves 98.9\% mean AUROC, 
surpassing the previous state-of-the-art method by a significant margin of 0.3 percentage points. 
The results indicate that custom diffusion processes crafted specifically for surface anomaly detection are a promising direction for future research.

\subsubsection{Acknowledgements}This work was in part supported by the ARIS research project L2-3169 (MV4.0), research programme P2-0214 and the supercomputing network SLING (ARNES, EuroHPC Vega).

%
%
\bibliographystyle{splncs04}
\bibliography{egbib}
\newpage
\setcounter{page}{1}
\title{TransFusion -- A Transparency-Based Diffusion Model for Anomaly Detection
\newline {\large Supplementary material}
}
\author{}
\authorrunning{M.~Fučka et al.}
\institute{}
\titlerunning{TransFusion}
\maketitle

In this supplementary material, we provide some details and supporting information that extend beyond the scope of the main manuscript and complement it well. Specifically, we show and discuss a number of failure cases, report detailed per-class localisation results, and examine the sensitivity of the proposed method to two hyperparameters. Then, we show the results obtained by finetuning hyperparameters to the individual classes, a practice which is used in many recent anomaly detection methods. Finally, we present a comprehensive collection of additional qualitative results, alongside a qualitative comparison with related work, to provide even more insights into the performance of the proposed method.

\section{Failure cases}

\begin{figure}[t]
    \centering
    \includegraphics[width=\columnwidth]{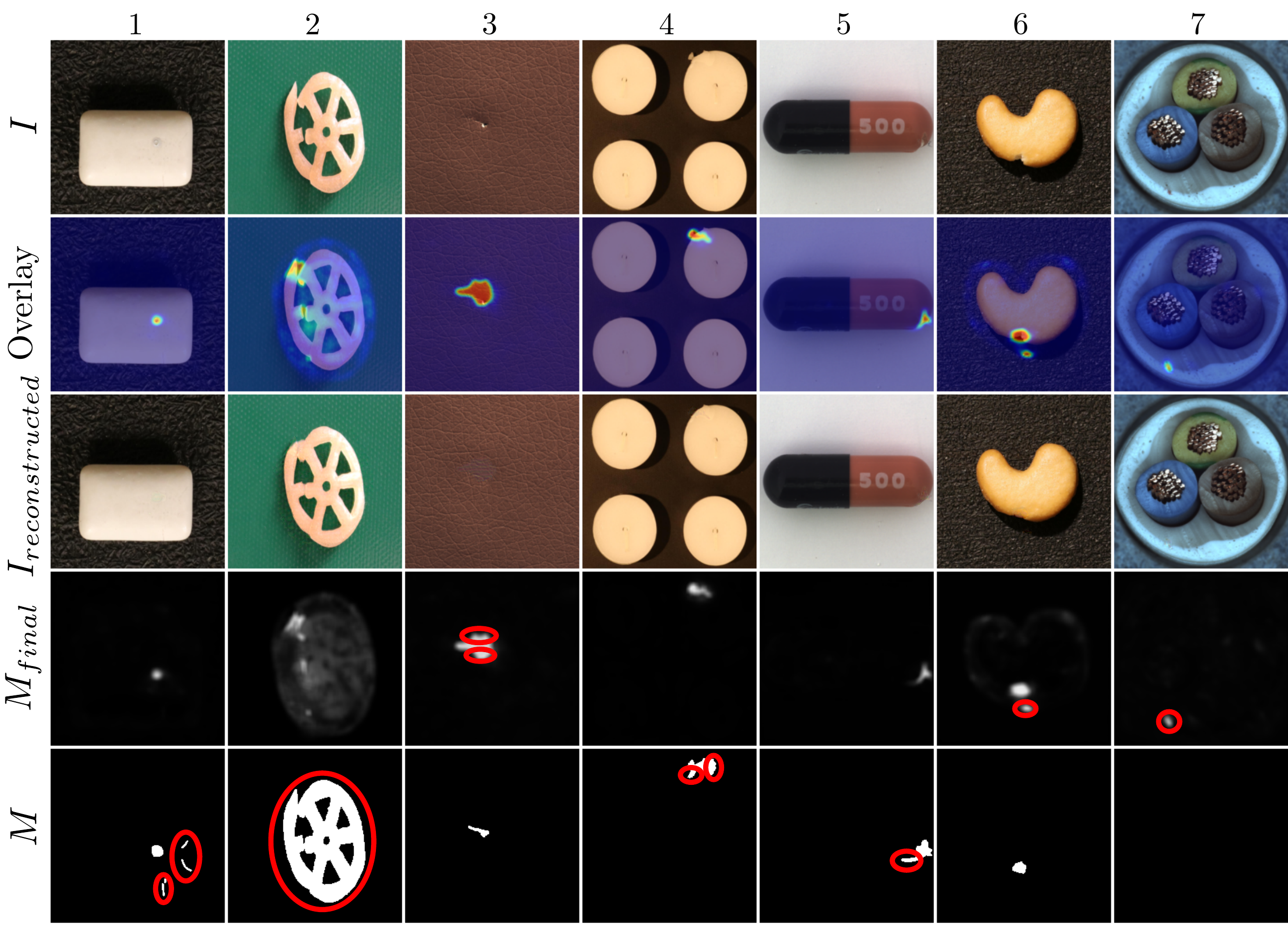}
    \caption{Failure case results. The anomalous images are shown in the first row, the overlay in the second row, the reconstructions in the third row, the predicted mask, and the real mask in the fourth and fifth rows, respectively. The biggest discrepancies between the predicted and ground truth masks are marked with \textcolor{red}{red circles}.
    }
    \label{fig:failure-cases}
\end{figure}

A few failure cases of TransFusion can be seen in Figure~\ref{fig:failure-cases}, where anomalies are not properly localized, or image regions are poorly reconstructed. TransFusion fails to segment tiny anomalous details (Column 1), and outputs masks that do not fit the ground truth in cases (Columns 2 to 5) where it is ambiguous what to annotate as the ground truth. For instance, in Column 2, TransFusion recognizes where the object broke, but the annotators annotate the whole object as anomalous. A similar thing can be noted in Column 3, where the annotators only annotated the hole while the leather around it is curved due to it, which could also be annotated as an anomaly. It also restores the normality of image regions that are relatively out of distribution but are not annotated (Columns 6 and 7). Some of these failure cases impact the anomaly localization score on VisA, where the anomaly masks are small and precise. MVTec AD contains larger anomalies. Therefore, the effect on the anomaly localization score is not as severe. However, the anomaly detection score is impacted.

\section{Per-class localization results}

Per-class localization results are provided in Table~\ref{tb:visa_loc_results} and in Table~\ref{tb:mvtec_loc_results}. The lowest scores are achieved for the Fryum and Cashew categories on VisA and for the Transistor and Cable categories on MVTecAD. We hypothesize that this is partly caused by the ambiguous anomalous regions that are difficult to annotate and common in these categories. A few of these ambiguous ground truths can be seen in Figure~\ref{fig:failure-cases}, more specifically in Rows 2, 3, 6 and 7. For instance, in Row 6, TransFusion also reconstructs a part of the shadow that is missing in the original image due to a crack in the cashew. Another example can be seen in Row 7, where TransFusion fixes a poke in the plastic around the cable. If there are multiple of them in the image this is considered an anomaly in the test set, so the annotation of this image is ambiguous.

\begin{table*}[h]
\centering
\setlength{\tabcolsep}{3pt}
\resizebox{\textwidth}{!}{
\begin{tabular}{lccccccccccccc}
                  \toprule \textbf{Category} & Candle & Capsules & Cashew & Chewing gum & Fryum & Macaroni1 & Macaroni2 & PCB1 & PCB2 & PCB3 & PCB4 & Pipe fryum & \textit{Average}\\
\midrule TransFusion & 88.6 & 97.3 & 82.8 & 83.2 & 77.8 & 94.0 & 95.6 & 92.4 & 85.1 & 92.0 & 89.4 & 87.9 & 88.8\\ \bottomrule
\end{tabular}
}
\caption{Detailed results for Transfusion for anomaly localization on VisA. All results are reported in AUPRO.}
\label{tb:visa_loc_results}
\end{table*}

\begin{table*}[h]
\centering
\setlength{\tabcolsep}{3pt}
\resizebox{\textwidth}{!}{
\begin{tabular}{lcccccccccccccccc}
                  \toprule \textbf{Category} & Carpet & Grid & Leather & Tile & Wood & Bottle & Cable & Capsule & Hazelnut & Metal nut & Pill & Screw & Toothbrush & Transistor & Zipper & \textit{Average}\\
\midrule TransFusion & 95.9 & 98.0 & 96.2 & 95.0 & 94.8 & 97.3 & 85.5 & 92.1 & 97.7 & 94.1 & 96.2 & 97.0 & 94.1 & 83.9 & 97.2 & 94.3 \\ \bottomrule
\end{tabular}
}
\caption{Detailed results for Transfusion for anomaly localization on MVTec AD. All results are reported in AUPRO.}
\label{tb:mvtec_loc_results}
\end{table*}

\section{Additional ablation study results}\label{ch:ablation_extra}

\begin{figure}[t]
    \centering
    \includegraphics[width=\columnwidth]{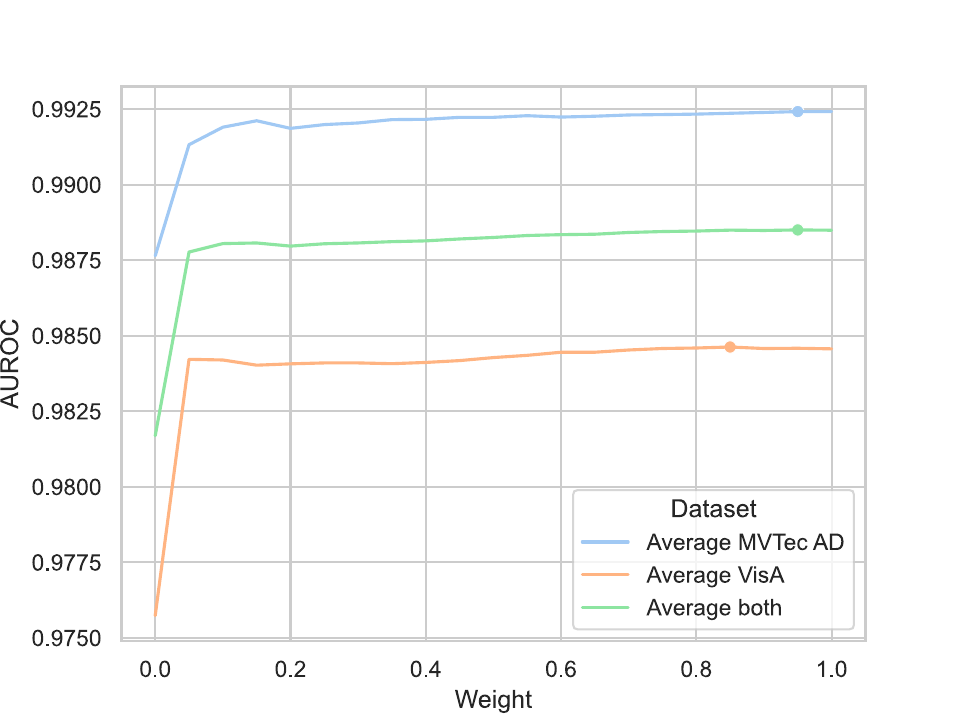}
    \caption{Average AUROC for different weights $\lambda$ in the final mask calculation. The maximum point of each line is represented with a dot.
    }

    \label{fig:weight}
\end{figure}

\noindent\textbf{Weight size.} In the final mask calculation (Eq.~(\ref{eq:final_mask_calc})) the weight $\lambda$ defines the impact of $M_{disc}$ and $M_{recon}$ on the final mask. TransFusion's performance under various $\lambda$ values is shown in Figure~\ref{fig:weight}. The results are robust for larger $\lambda$ values, where $M_{disc}$ has a higher impact on the final mask. However, the best results are achieved with $\lambda$ values at which $M_{recon}$ still impacts the final mask.

\begin{figure}[t]
    \centering
    \includegraphics[width=\columnwidth]{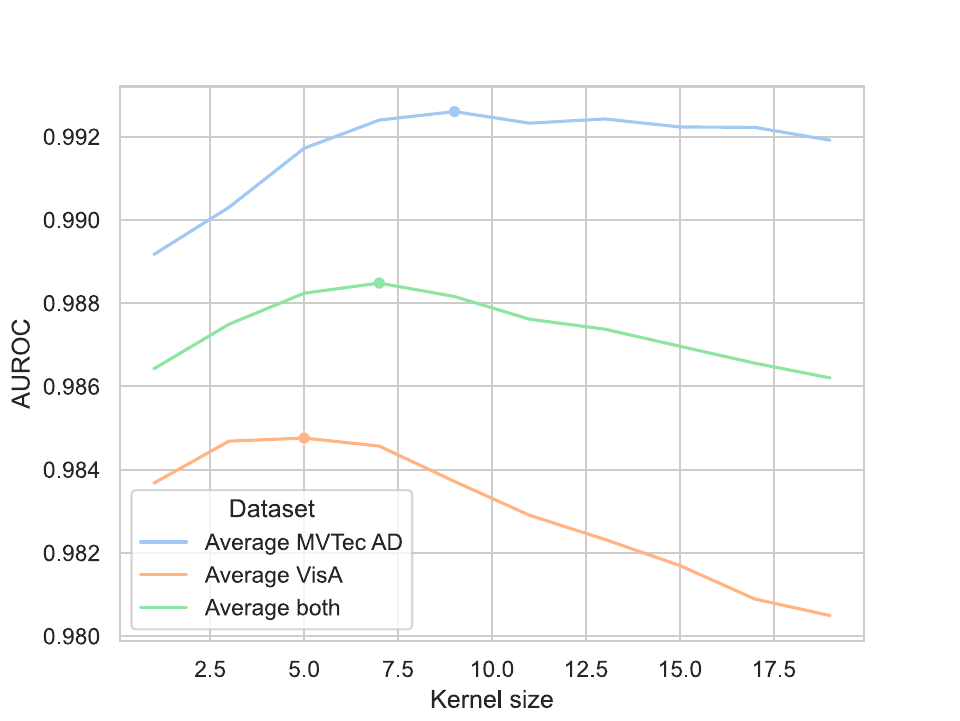}
    \caption{Average AUROC for different kernel sizes $n$ in the final mask calculation. The maximum point of each line is represented with a dot.
    }

    \label{fig:kernel}
\end{figure}

\noindent\textbf{Kernel size.} To determine the final mask calculation as described in Eq.~(\ref{eq:final_mask_calc}), we incorporated a mean filter $f_n$ of size $n$ into the formulation. Here we explore TransFusion's behaviour under various values of $n$. The results can be seen in Figure~\ref{fig:kernel}. Note that higher values of $n$ quickly deteriorate the performance on the VisA dataset due to the scale of anomalies present in the dataset. On the MVTec AD dataset, high kernel sizes have little to no effect on the anomaly detection performance.

\section{Per-class tuned results for anomaly detection}

Some recent works~\cite{ddpm-noise-anomaly, cflow-ad, simplenet} report performance where hyperparameter tuning was done for each class individually. We maintain a single set of hyperparameters for all experiments in the paper. For instance, the total number of epochs was set in stone, and the result was calculated using the model from the final epoch. For the sake of completeness, we report results where the total number of epochs was optimized for each class. These results enable future works to be compared with per-class tuned models. The results are shown in Table~\ref{tb:visa_det_results_prophet} and Table~\ref{tb:mvtec_det_results_prophet}. Results on VisA~\cite{visa} exceed the current highest score by $0.9$\%, and results on  MVTec AD~\cite{mvtec} improve even further.

\begin{table*}[t]
\centering
\setlength{\tabcolsep}{3pt}
\resizebox{\textwidth}{!}{
\begin{tabular}{lccccccccccccc}
                  \toprule \textbf{Category} & Candle & Capsules & Cashew & Chewing gum & Fryum & Macaroni1 & Macaroni2 & PCB1 & PCB2 & PCB3 & PCB4 & Pipe fryum & \textit{Average}\\
\midrule Detection & 98.3 & 99.7 & 96.8 & 99.9 & 98.7 & 99.4 & 96.8 & 99.1 & 99.9 & 99.5 & 99.6 & 99.8 & 99.0 \\
\bottomrule
\end{tabular}
}
\caption{Best possible results for TransFusion when we choose the optimal number of epochs for each class on VisA. Anomaly detection results are reported in AUROC.}
\label{tb:visa_det_results_prophet}
\end{table*}

\begin{table*}[t]
\centering
\setlength{\tabcolsep}{3pt}
\resizebox{\textwidth}{!}{
\begin{tabular}{lcccccccccccccccc}
                  \toprule \textbf{Category} & Carpet & Grid & Leather & Tile & Wood & Bottle & Cable & Capsule & Hazelnut & Metal nut & Pill & Screw & Toothbrush & Transistor & Zipper & \textit{Average}\\
\midrule Detection & 99.8 & 100 & 100 & 100 & 99.9 & 100 & 98.4 & 98.8 & 100 & 100 & 99.5 & 97.2 & 100 & 98.8 & 100 & 99.5\\
\bottomrule
\end{tabular}
}
\caption{Best possible results for TransFusion when we choose the optimal number of epochs for each class on MVTec AD. Anomaly detection results are reported in AUROC.}
\label{tb:mvtec_det_results_prophet}
\end{table*}

\section{Additional qualitative results}\label{ch:qualitative_extra}

\begin{figure*}
    \centering
    \includegraphics[width=\textwidth]{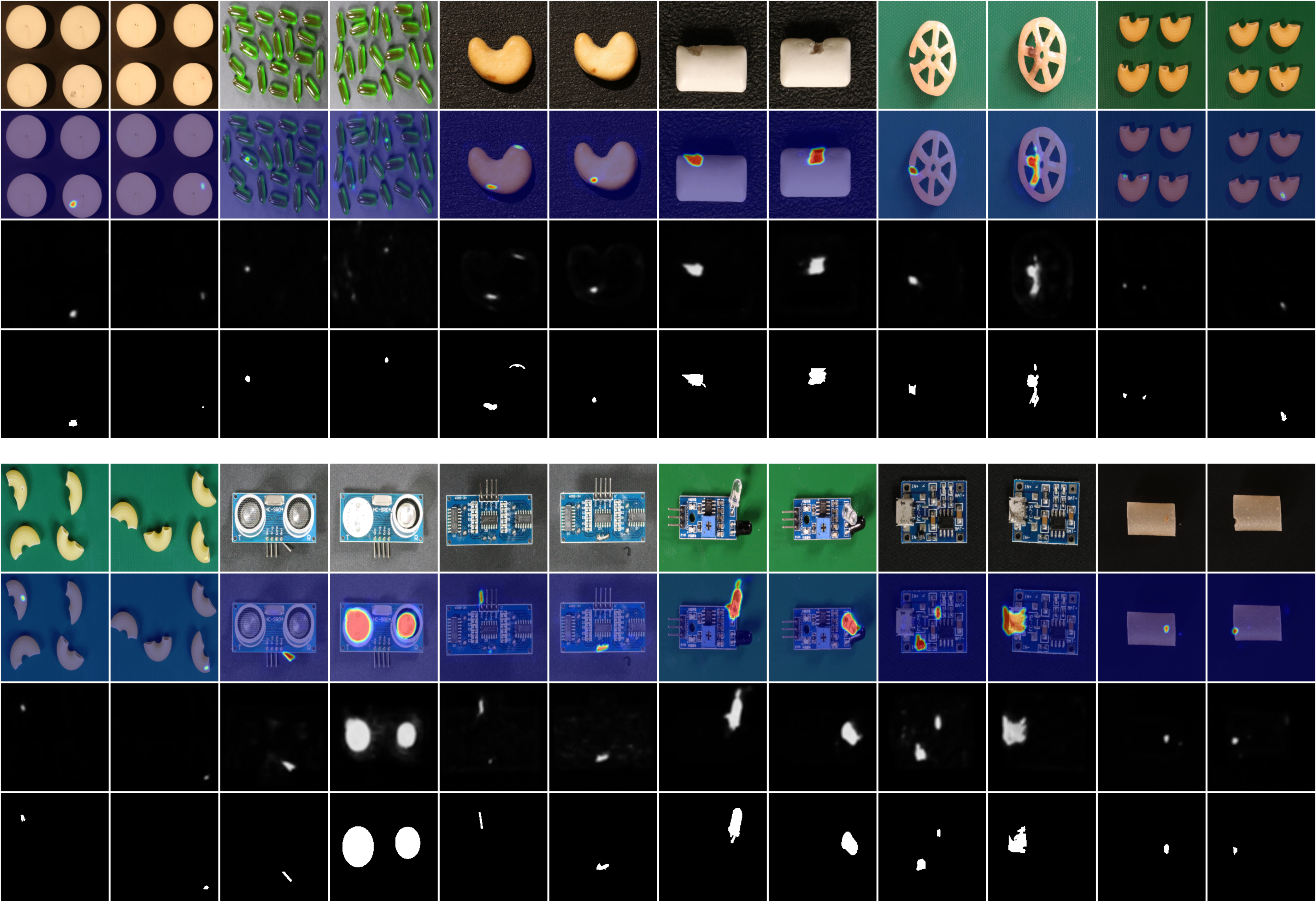}
    \caption{Qualitative examples on VisA dataset. The original image, the anomaly map overlay, the anomaly map and the ground truth map are shown.}
    \label{fig:qualitative_visa}
\end{figure*}

\begin{figure*}
    \centering
    \includegraphics[width=\textwidth]{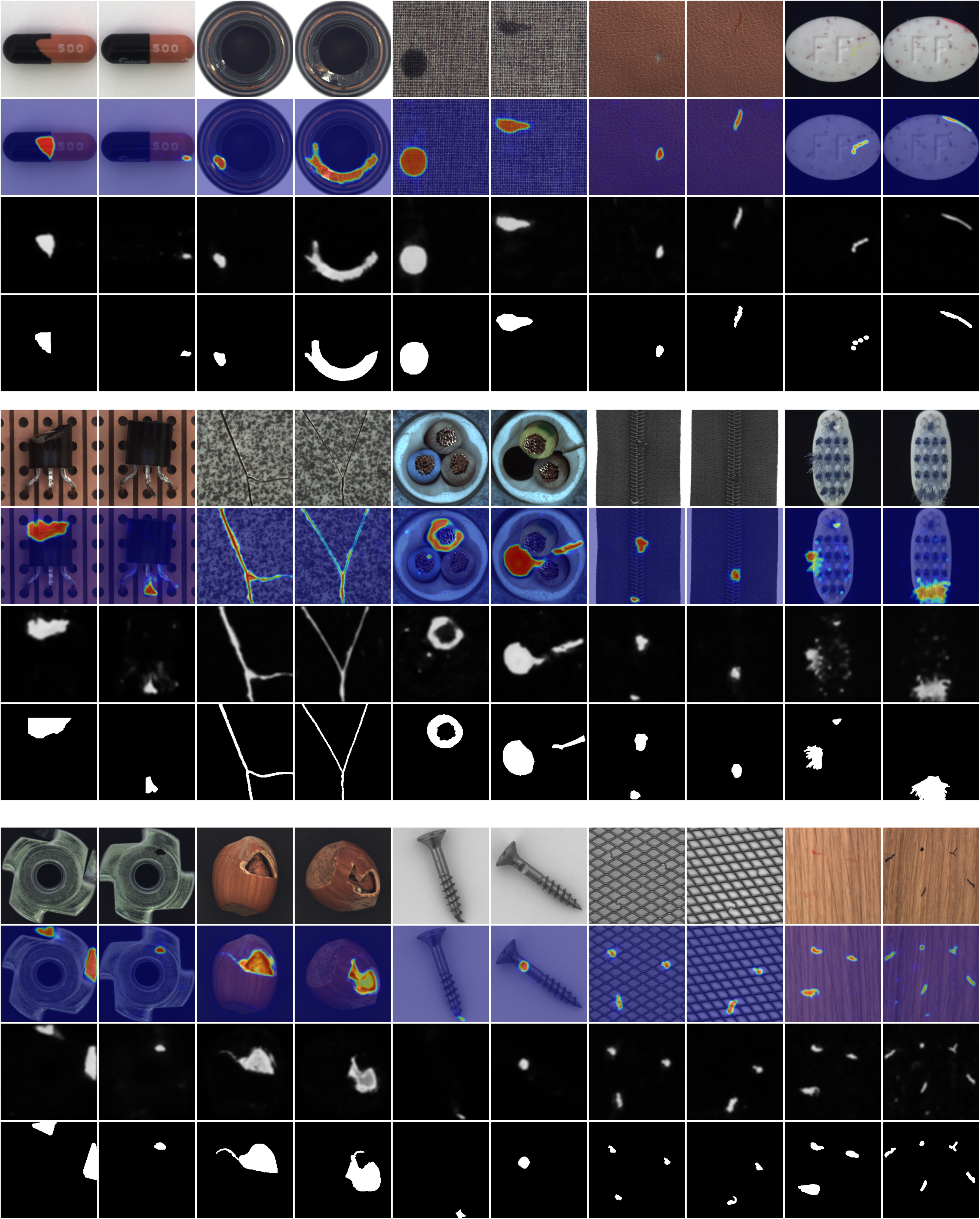}
    \caption{Qualitative examples on MVTec AD dataset. The original image, the anomaly map overlay, the anomaly map and the ground truth map are shown.}
    \label{fig:qualitative_mvtec}
\end{figure*}

In this section, we provide more qualitative results. Figure~\ref{fig:qualitative_visa} and Figure~\ref{fig:qualitative_mvtec} show some result samples from each category on both datasets. As we can observe, TransFusion outputs very precise masks that closely match the ground truth annotation in the vast majority of cases.

\section{Additional qualitative comparisons to other methods}

\begin{figure*}
    \centering
    \includegraphics[width=0.7\textwidth]{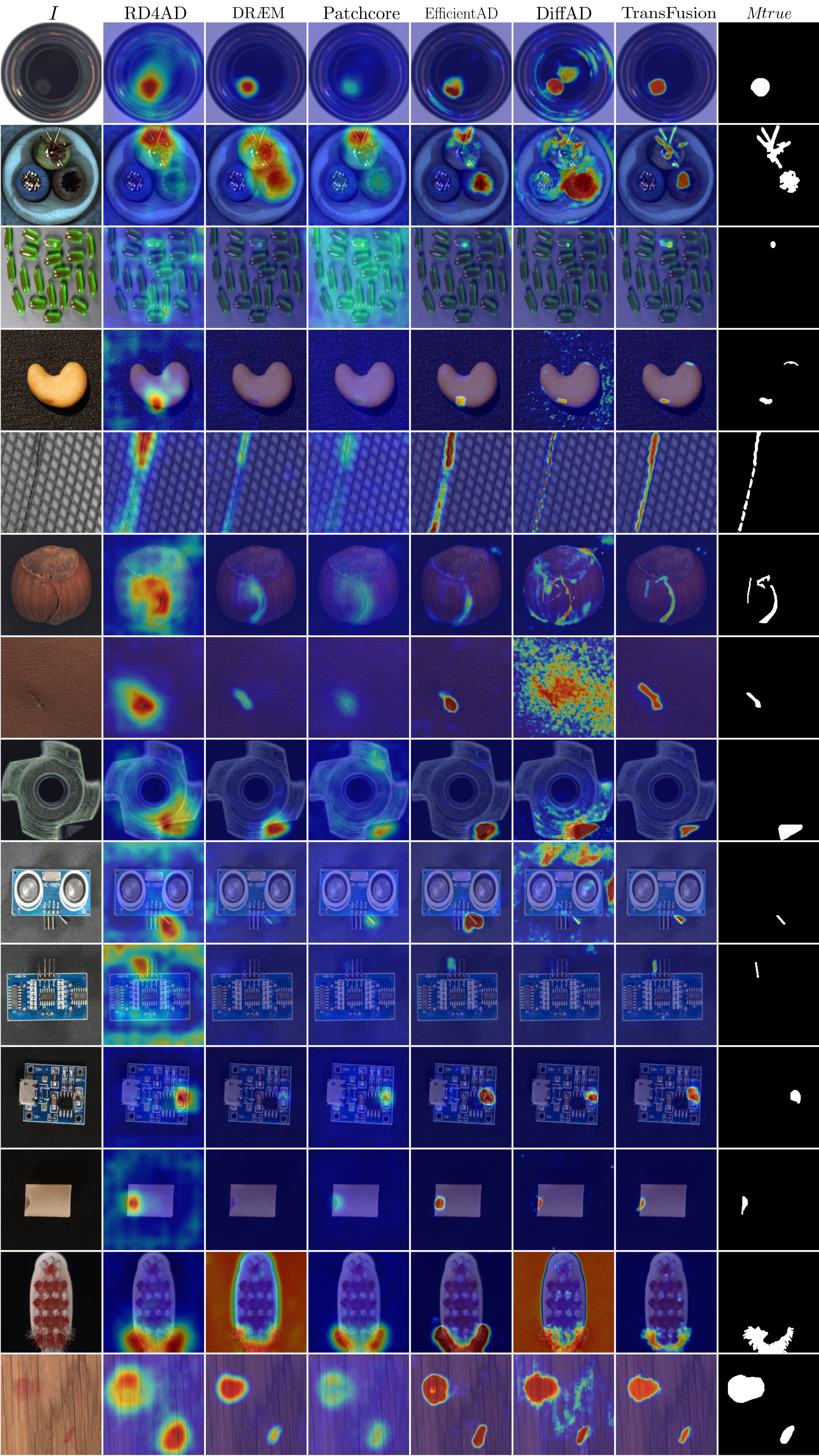}
    \caption{Qualitative comparison of the masks produced by TransFusion and five other state-of-the-art methods. The anomalous images are shown in the first column. The middle six columns show the anomaly mask generated by RD4AD~\cite{reverse_dist}, DR{\AE}M~\cite{draem}, Patchcore~\cite{patchcore}, EfficientAD~\cite{efficientad}, DiffAD~\cite{ldm_draem} and TransFusion respectively. The last column shows the ground truth anomaly mask.}
    \label{fig:mask_comp_supl}
\end{figure*}

This section provides more qualitative mask comparisons to other state-of-the-art methods. We compared TransFusion with DRAEM~\cite{draem}, RD4AD~\cite{reverse_dist}, Patchcore~\cite{patchcore} and DiffAD~\cite{ldm_draem}. The results can be seen in Figure~\ref{fig:mask_comp_supl}.

\end{document}